\documentclass[10pt,twocolumn,letterpaper]{article}

\usepackage{iccv}              
\usepackage{amsmath}
\usepackage{multirow}
\usepackage{tabularx}
\usepackage{amssymb}

\newcolumntype{R}[1]{>{\raggedleft\arraybackslash}p{#1}} 
\newcolumntype{L}[1]{>{\raggedright\arraybackslash}p{#1}} 
\newcolumntype{C}[1]{>{\centering\arraybackslash}p{#1}} 

\newcolumntype{Y}{>{\raggedright\arraybackslash}X}
\newcolumntype{Z}{>{\raggedleft\arraybackslash}X}
\newcolumntype{A}{>{\centering\arraybackslash}X}
\definecolor{iccvblue}{rgb}{0.21,0.49,0.74}
\usepackage[pagebackref,breaklinks,colorlinks,allcolors=iccvblue]{hyperref}

\def\paperID{11888} 
\def\confName{ICCV}
\def\confYear{2025}

\title{PersPose: 3D Human Pose Estimation with Perspective Encoding and Perspective Rotation}

\author{Xiaoyang Hao$^{*}$ \ \ \ Han Li$^{*\dagger}$\\
Southern University of Science and
Technology, China\\
{\tt\small \{haoxy2022, lih2022\}@mail.sustech.edu.cn}
}

\begin{document}
\maketitle
\renewcommand{\thefootnote}{\fnsymbol{footnote}}
\footnotetext[1]{Equal contribution}
\footnotetext[2]{Corresponding author}

\begin{abstract}
Monocular 3D human pose estimation (HPE) methods estimate the 3D positions of joints from individual images. 
Existing 3D HPE approaches often use the cropped image alone as input for their models. However, the relative depths of joints cannot be accurately estimated from cropped images without the corresponding camera intrinsics, which determine the perspective relationship between 3D objects and the cropped images. 
In this work, we introduce Perspective Encoding (PE) to encode the camera intrinsics of the cropped images. 
Moreover, since the human subject can appear anywhere within the original image, 
the perspective relationship between the 3D scene and the cropped image differs significantly, which complicates model fitting. 
Additionally, the further the human subject deviates from the image center, the greater the perspective distortions in the cropped image. 
To address these issues, we propose Perspective Rotation (PR), a transformation applied to the original image that centers the human subject, thereby reducing perspective distortions and alleviating the difficulty of model fitting.
By incorporating PE and PR, we propose a novel 3D HPE framework, PersPose. Experimental results demonstrate that PersPose achieves state-of-the-art (SOTA) performance on the 3DPW, MPI-INF-3DHP, and Human3.6M datasets. For example, on the in-the-wild dataset 3DPW, PersPose achieves an MPJPE of 60.1 mm, 7.54\% lower than the previous SOTA approach. 
Code is available at: \url{https://github.com/KenAdamsJoseph/PersPose}.
\end{abstract}    
\section{Introduction}
\label{sec:intro}

\begin{figure}[t]
  \centering
   \includegraphics[width=1.0\linewidth]{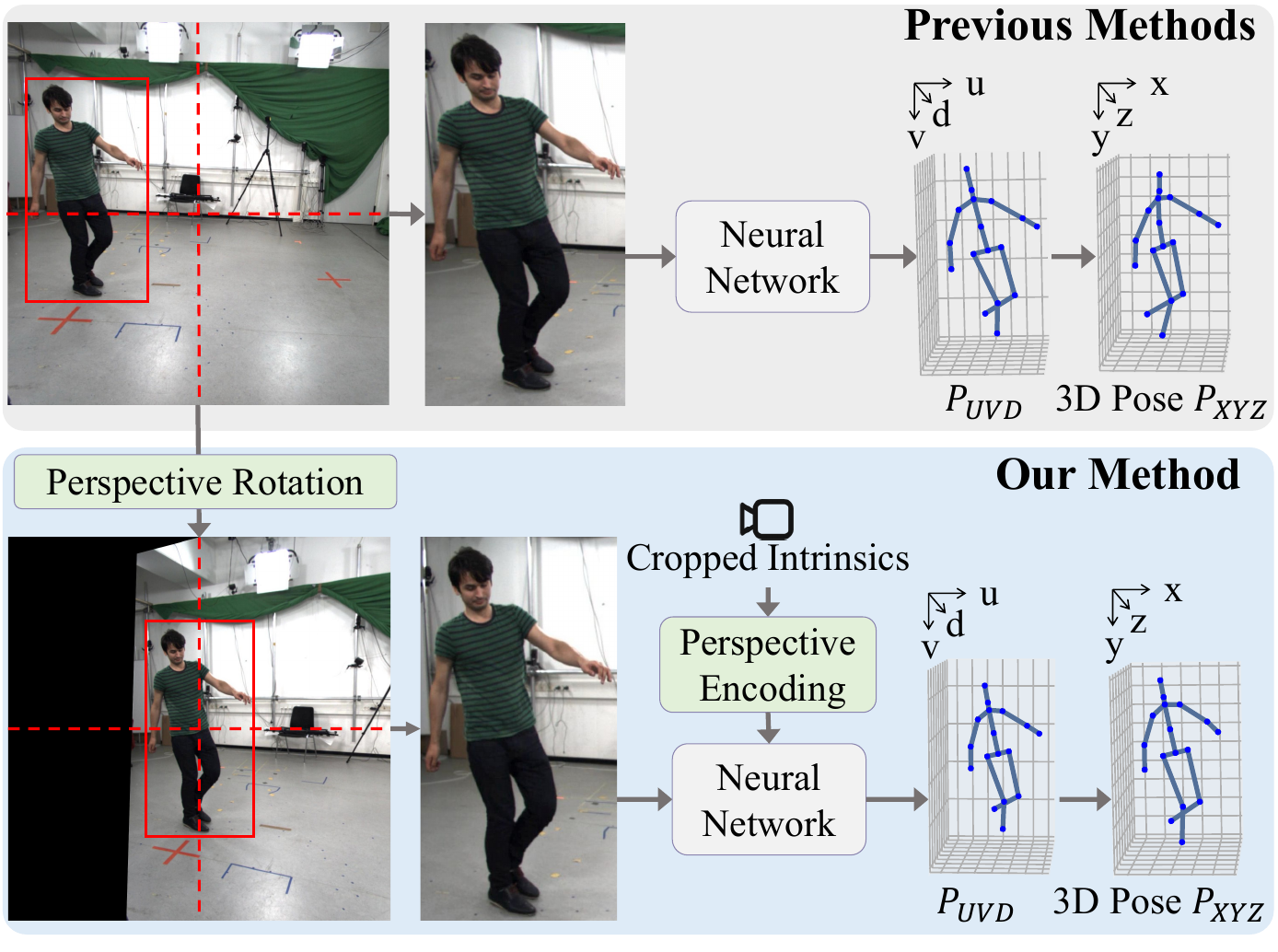}
   \vspace{-0.6cm}
   \caption{Comparison of previous 3D HPE frameworks and ours.}
   \vspace{-0.3cm}
   \label{fig:intro}
\end{figure}

Estimating 3D human pose from a single image is a crucial task in computer vision, with numerous practical applications such as AR/VR and human-computer interaction. The objective of 3D human pose estimation (HPE) is to recover the 3D coordinates of human joints from images, whereas 3D human pose and shape estimation (HPS) aims to predict the 3D coordinates of human mesh vertices. Some 3D HPS methods~\cite{HybrIK2021cvpr,ma20233d,shetty2023pliks,zhang2023ikol,fang2023learning} treat 3D HPE as a sub-task, first estimating the 3D positions of the joints (or a larger set of mesh vertices) and then reconstructing the human mesh.

Existing 3D HPE methods~\cite{sun2018integral,ma2021context,wang2020predicting,li2021human,pavlakos2017coarse} and 3D HPS methods~\cite{HybrIK2021cvpr,ma20233d,shetty2023pliks,zhang2023ikol,fang2023learning,li2023niki} typically use cropped images as input to estimate 3D joint positions, meaning that the region corresponding to the human bounding box is cropped from the original image, as illustrated in Figure~\ref{fig:intro}. However, the CLIFF method~\cite{li2022cliff} demonstrates that using cropped images alone cannot accurately estimate global rotation (i.e., the orientation of the human in the camera coordinate system), 
thus it employs both bounding box information and the cropped images as inputs to the neural network. 
In this paper, we demonstrate through the example in Figure \ref{fig:crop1} that when only the cropped image is provided as input, the relative depths of joints cannot be accurately estimated either. 
Additionally, the example in Figure \ref{fig:crop2} reveals that even when the full image is used, the absence of field of view angle (FOV) information results in an inaccurate estimation of the joints' relative depths. 
Moreover, cropping can be considered as a modification of the camera intrinsics and sensor resolution; in this context, the camera intrinsics corresponding to the cropped image effectively encapsulate both the cropping and FOV information. 
We refer to the camera intrinsics corresponding to the cropped image as cropped intrinsics, denoted as \(K^{\text{crop}}\), which include the focal length \(f^{\text{crop}}\) and the principal point coordinates \(c_x^{\text{crop}}\) and \(c_y^{\text{crop}}\). 
We propose a Perspective Encoding (PE) module that encodes the cropped intrinsics as a 2D PE map, which is then jointly fed into the CNN with the cropped image.

Since a person can appear at any location within an image, the principal points of the cropped intrinsics $c_x^{\text{crop}}$ and $c_y^{\text{crop}}$ vary significantly across different samples. 
This reflects substantial changes in the perspective relationship between the 3D scene and the cropped image. 
Additionally, the further the human subject deviates from the image center, the more pronounced the perspective distortions of the cropped image become. 
These factors increase the difficulty of model fitting. 
To mitigate these issues, we propose a Perspective Rotation (PR) module, which centers the human subject and generates a centered image from the original image, thereby ensuring that $c_x^{\text{crop}}$ and $c_y^{\text{crop}}$ remain fixed across samples while also reducing perspective distortions.

In this study, we propose a 3D HPE framework, PersPose, which incorporates the proposed PE and PR modules, as illustrated in Figure \ref{fig:intro}. 
First, the original image undergoes the PR module, yielding a centered image (see Section~\ref{sec:pr} for details), which is then cropped from the center. 
Subsequently, the cropped intrinsics $K^{\text{crop}}$, calculated using Eq. \ref{eq:K_crop}, are encoded by the PE module to form a PE map, as detailed in Section~\ref{sec:encoding}. 
The PE map and the cropped image are then fed into a CNN backbone,  and then a decoder is used to predict the 3D Pose $P_{\text{XYZ}}$. 
The contributions of this paper can be summarized as follows: 
\begin{itemize}
\item 
Without camera intrinsics information, the relative depths of joints cannot be accurately estimated. To mitigate this previously overlooked limitation, we propose an innovative PE module that encodes the cropped intrinsics as a 2D PE map, which is then jointly fed into the CNN with the cropped image. 
\item Due to the variation in a person's location within the original images, the perspective relationships between the 3D scene and the cropped image vary significantly, which complicates model fitting. To address this, we propose a novel PR module that centers the human subject, thereby reducing perspective distortions and alleviating the difficulty of model fitting.
\item We propose a 3D HPE framework, PersPose, incorporating our PE and PR modules. To evaluate our framework, we conduct comprehensive experiments on 3DPW~\cite{3DPW2018}, Human3.6M~\cite{human36M2013}, and MPI-INF-3DHP~\cite{mehta2017monocular} datasets. Experimental results show that our framework outperforms existing state-of-the-art (SOTA) methods. 
Ablation studies further validate the effectiveness of the proposed PE and PR modules. 
\end{itemize}

\section{Related Work}
\label{sec:related}
\nocite{you2023co,patel2021agora}

Some 3D HPE methods~\cite{pavlakos2017coarse,sun2018integral,ma2021context,li2021human} directly estimate 3D joint coordinates. \citet{pavlakos2017coarse} proposed to voxelize the 3D space into a 3D probabilistic heatmap.
\citet{sun2018integral} proposed a differentiable soft-argmax operation to derive joint coordinates from estimated heatmaps. 
\citet{ma2021context} proposed an attention-based context modeling framework that propagates structural cues across joints and effectively suppresses absurd 3D pose estimates. 
\citet{li2021human} designed RLE from a maximum‑likelihood perspective to learn flexible output distributions for joint positions, thereby facilitating the training process.

Other methods~\cite{wang2020predicting,HybrIK2021cvpr,ma20233d,shetty2023pliks}, including ours, first estimate 2D pixel coordinates and relative depths of joints, and then convert these into 3D joint coordinates based on camera intrinsics. 
This approach facilitates joint training with 2D keypoint datasets.
The principal point is usually near the image center. As for focal length, some studies assume it to be unknown. 
Some methods~\cite{kanazawa2018hmr,MuhammedKocabas2021PAREPA} employ a weak perspective model, which effectively corresponds to using a large fixed focal length. \citet{kissos2020beyond} proposed estimating a rough focal length using image resolution. 
The SPEC method~\cite{kocabas2021spec} incorporates a camera calibration sub-network to estimate the focal length directly from images, while the Zolly method~\cite{wang2023zolly} estimates the human-to-camera distance and scale from images to derive the focal length. 
However, the focal length of a camera is usually obtainable (e.g., through device APIs/SDKs) and is essential for estimating the relative depths of joints, as illustrated in Figure~\ref{fig:crop2}. 

The CLIFF method~\cite{li2022cliff} demonstrates that global rotation, i.e., the orientation of the human in the camera coordinate system, cannot be accurately inferred from cropped images alone. Their work underscores the necessity of incorporating bounding box information. 
To address the adverse effects of cropping on global orientation, \citet{yao2024rotated} proposed applying a compensatory rotation within their self-supervised 3D HPE framework to correct the global orientation of the estimated 3D skeleton. 

Furthermore, some 3D HPE methods utilize 2D keypoint sequences as input. The recent Ray3D method~\cite{zhan2022ray3d} leverages camera intrinsics to convert 2D keypoints into 3D rays, and subsequently employs a temporal network to recover the 3D pose sequence from the 3D ray sequence. 
Additionally, methods like ElePose~\cite{wandt2022elepose} and EPOCH~\cite{garau2024epoch} estimate camera pose within unsupervised 3D HPE frameworks.

\section{Importance of Camera Intrinsics in 3D HPE}
\label{sec:importance}
In the following, we first introduce camera intrinsics and FOV. Then, we demonstrate the limitation of using cropped images alone for 3D HPE. Finally, we discuss the importance of FOV for 3D HPE when images are not cropped. 
Since the cropping operation can be considered as a modification of the camera intrinsics and resolution, the cropped intrinsics are inherently linked to the cropping operation and FOV; this underscores the significance of camera intrinsics in 3D HPE. 

\paragraph{Camera Intrinsics and FOV.}
Camera intrinsics are used to map 3D coordinates onto 2D image coordinates, playing a pivotal role in 3D HPE. The intrinsic matrix $K$ is a $3 \times 3$ matrix: 
\begin{equation}
\label{eq:K}
K = \begin{bmatrix}
f & 0 & c_x \\
0 & f & c_y \\
0 & 0 & 1
\end{bmatrix} ,
\end{equation}
where $f$ denotes the effective focal length in pixel units. If the virtual sensor\footnote{The virtual sensor is a mathematical abstraction positioned on the virtual image plane, located at a distance equal to the focal length in front of the optical center.} moves along the $z$-axis, $f$ scales accordingly. The coordinates $(c_x, c_y)$ represent the principal point, which is typically located near the image center. Adjusting $(c_x, c_y)$ is equivalent to moving the virtual sensor within the $x$-$y$ plane.

The FOV quantifies the extent of the observable world through a camera. 
FOV is determined by the effective focal length $f$ and the image size:
\begin{align}
\text{FOV}_x = 2 \arctan\left(\frac{w}{2f}\right), 
\text{FOV}_y = 2 \arctan\left(\frac{h}{2f}\right), 
\end{align}
where $\text{FOV}_x$ and $\text{FOV}_y$ are the horizontal and vertical fields of view, respectively. $h$ and $w$ represent the height and width of the captured image in pixels.

\paragraph{Limitation of Depth Estimation from Cropped Images.}

\begin{figure}[t]
  \centering
   \includegraphics[width=1.0\linewidth]{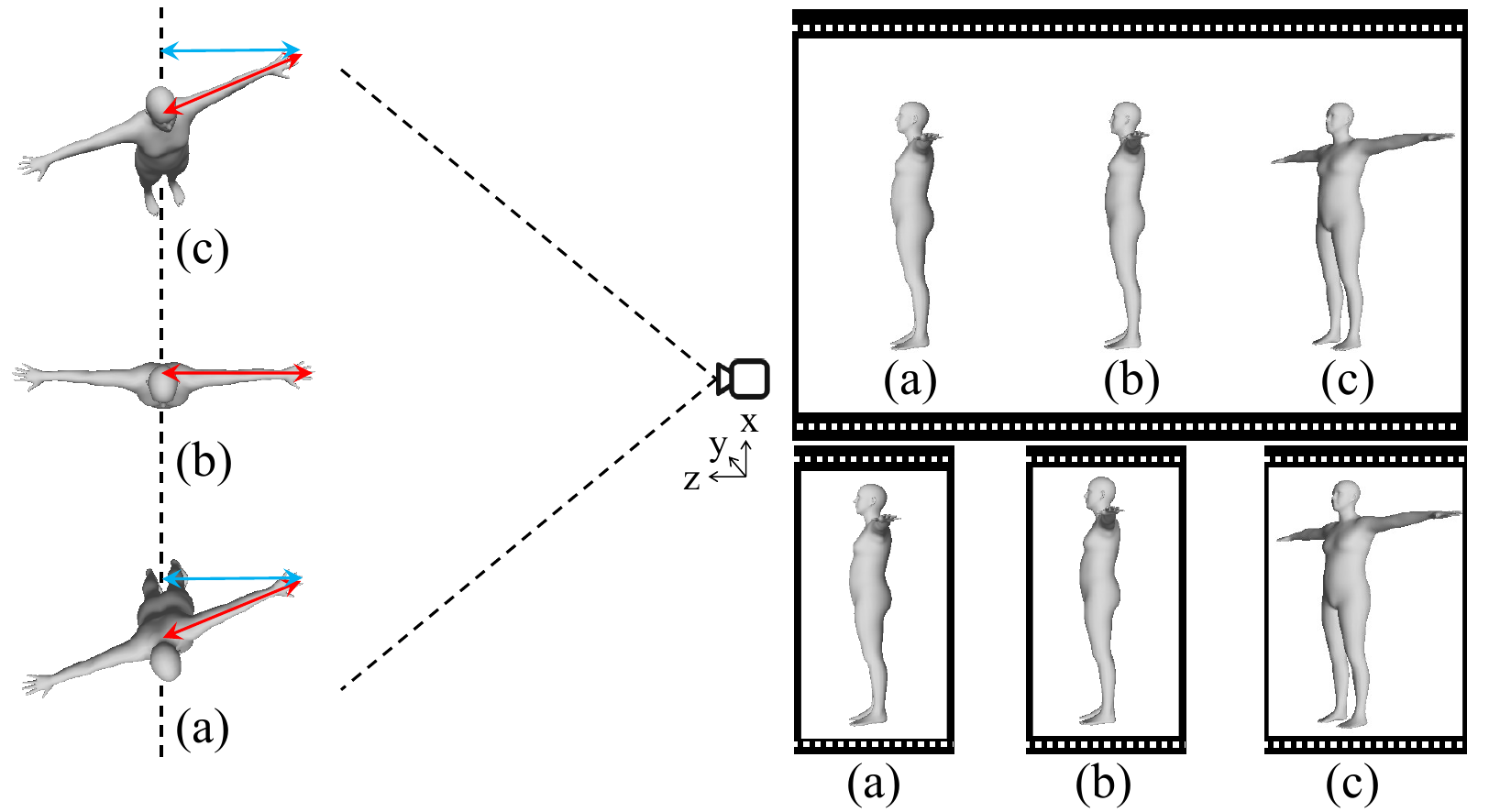}
   \vspace{-0.4cm}
   \caption{\textbf{Examples that illustrate the limitation of depth estimation using cropped images alone.} A single camera captures three human subjects with the identical body shape, and the resulting images are subsequently cropped. The blue lines denote the relative depths (i.e., to the pelvis) of a particular joint, while the red lines measure the arm lengths. In addition, the red lines for (a) and (c) are parallel. For cropped images, two key observations are noted: 1) subjects (a) and (b), despite having different relative depths, yield the same cropped images; 2) subjects (a) and (c), which share the same relative depths, produce distinct cropped images.}
   \vspace{-0.2cm}
   \label{fig:crop1}
\end{figure}

\begin{figure}[t]
  \centering
   \includegraphics[width=0.8\linewidth]{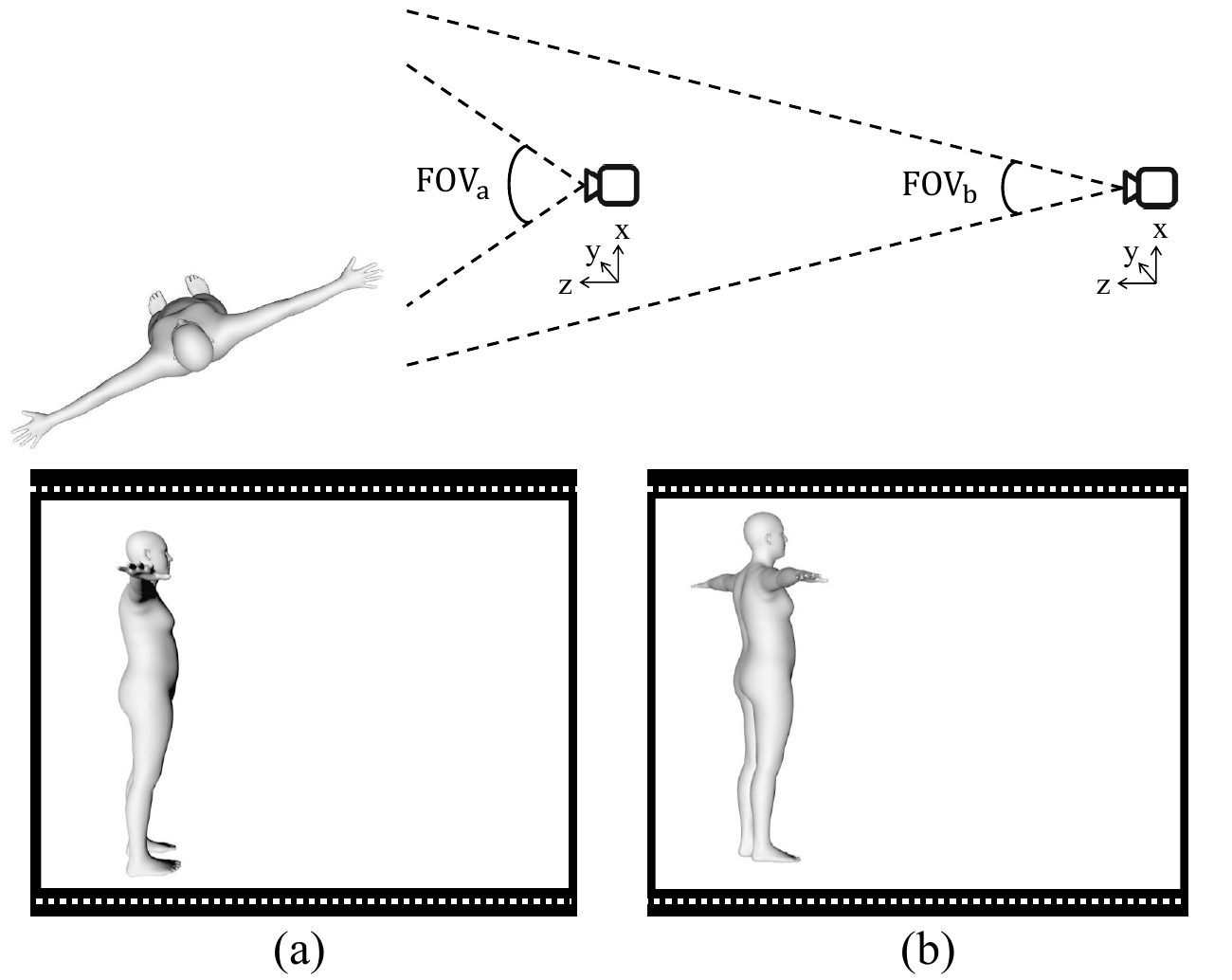}
   \caption{\textbf{Examples that illustrate the importance of FOV to depth estimation.} Two cameras at varying distances from a human subject capture images, resulting in images (a) and (b). Both cameras have the same optical axes but different FOV. Although there are significant visual differences between the two images, they share identical relative depth labels.}
   \vspace{-0.2cm}
   \label{fig:crop2}
\end{figure}

In the preprocessing stage of existing 3D HPE methods~\cite{sun2018integral,ma2021context,wang2020predicting,li2021human} and 3D HPS methods~\cite{HybrIK2021cvpr,ma20233d,shetty2023pliks,zhang2023ikol,fang2023learning}, original images are typically cropped based on the bounding boxes of human subjects. However, using only cropped images for 3D HPE introduces significant difficulty due to the lack of important information about the position of the human subject within the camera's view frustum. This information directly relates to how human subjects are projected onto images from the camera's perspective. 

Figure~\ref{fig:crop1} gives visual examples demonstrating the limitation of using only cropped images for 3D HPE. A camera captures three human subjects. 
The original image is displayed in the upper right, while the lower right shows three cropped images corresponding to the three human subjects.
For subjects (a) and (b), their cropped images are very similar despite having different relative depth labels. Similarly, the cropped images of subjects (a) and (c) are quite different, even though they share the same relative depth labels. 

Furthermore, cropping an image is equivalent to capturing a photo using another camera with the same camera extrinsics but with different camera intrinsics and resolution. The same camera extrinsics mean keeping the position of the pinhole and the camera's orientation fixed in the world coordinate system, and the modification of the camera intrinsics from $K$ to $K^{\text{crop}}$ can be geometrically viewed as a translation of the virtual sensor.
Specifically, $K^{\text{crop}}$ is calculated by: $K^{\text{crop}} = A K$, 
where $A$ is the affine transformation matrix corresponding to the cropping operation. 
The matrix $A$ is defined as
\begin{equation}
A = \begin{bmatrix}
s & 0 & t_u \\
0 & s & t_v \\
0 & 0 & 1
\end{bmatrix},
\end{equation}
where $s = w^{\text{bbox}}/w^{\text{crop}}$ (or $s = h^{\text{bbox}}/h^{\text{crop}}$), $
t_u = w^{\text{crop}}/2 - s\, c_u^\text{bbox}, \quad t_v = h^{\text{crop}}/2 - s\, c_v^\text{bbox}$.
Here, $w^{\text{crop}}$ and $h^{\text{crop}}$ denote the resolution of the cropped image, $w^{\text{bbox}}$ and $h^{\text{bbox}}$ represent the bounding box resolution, and $c_u^{\text{bbox}}$ and $c_v^{\text{bbox}}$ are the center coordinates of the bounding box.
Given the original camera intrinsics in Eq. \ref{eq:K},
the cropped intrinsics $K^{\text{crop}}$ become
\begin{equation}
\label{eq:K_crop}
K^{\text{crop}} = A K = 
\begin{bmatrix}
f^{\text{crop}} & 0 & c_x^{\text{crop}} \\
0 & f^{\text{crop}} & c_y^{\text{crop}} \\
0 & 0 & 1
\end{bmatrix},
\end{equation}
where $f^{\text{crop}}=s f$, $c_x^{\text{crop}}=s c_x + t_u$, and $c_y^{\text{crop}}=s c_y + t_v$.

Consequently, the cropped intrinsics $K^{\text{crop}}$ can be utilized to map 3D coordinates onto the 2D coordinate system of the cropped image.

\paragraph{Importance of FOV to Depth Estimation.} 

In Figure \ref{fig:crop2}, two cameras with the same optical axis capture the same human subject. The left camera has a larger field of view angle, i.e., $FOV_a > FOV_b$, and it is closer to the human subject compared to the right camera. 
These two cameras produce images (a) and (b). 
Note that (a) and (b) look quite different: for image (a), the depth of the right wrist relative to the pelvis seems larger than in image (b).
However, images (a) and (b) actually have the same relative depth labels. This example demonstrates that the FOV must also be considered to infer relative depths accurately.

\section{Method}

\begin{figure*}[t]
  \centering
   \includegraphics[width=1.\linewidth]{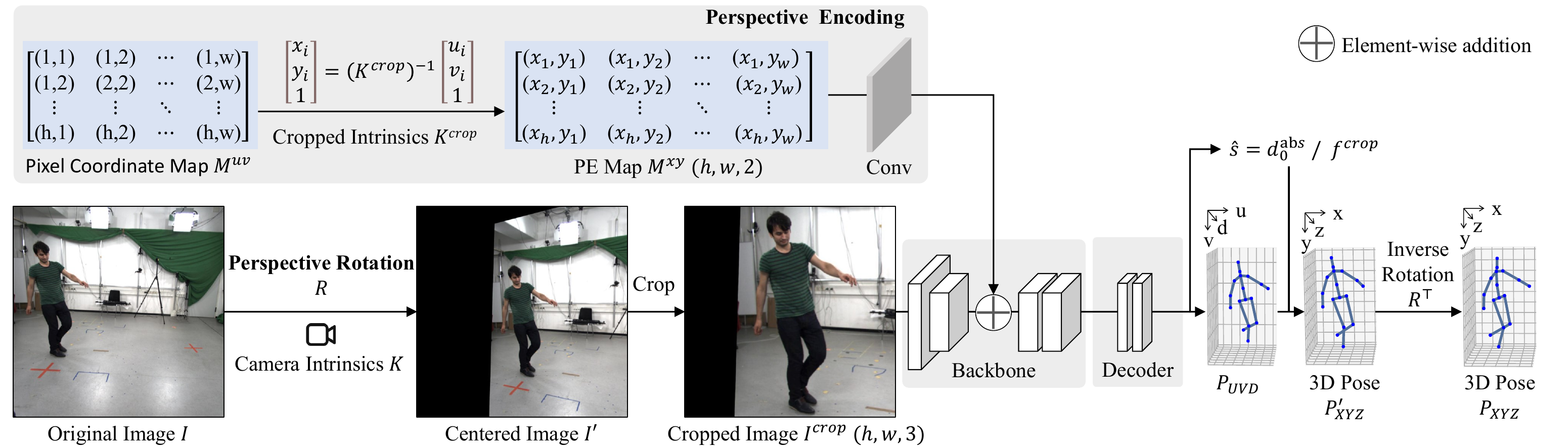}
   \caption{\textbf{Overview of the PersPose.} PersPose introduces two key components: 1) PE, which encodes the cropped intrinsics as a 2D map $M^{xy}$; and 2) PR, which centers the human subject in the image, reducing the difficulty of model fitting.} 
   \label{fig:pipeline}
\end{figure*}

\begin{figure}[t]
\centering
\includegraphics[width=0.8\linewidth]{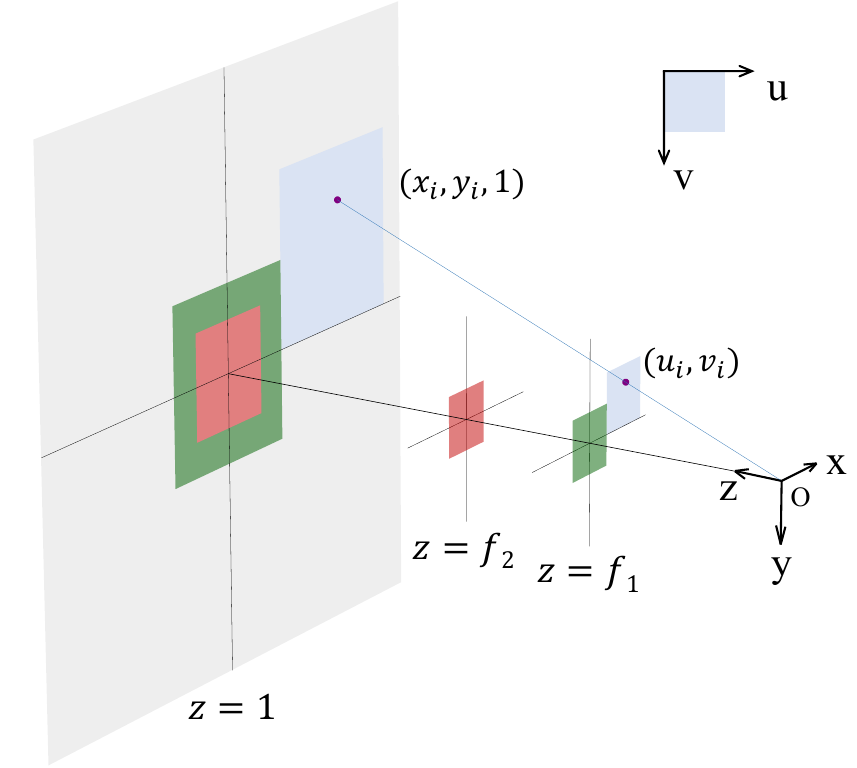}
\caption{\textbf{Perspective Encoding}. 
The grey region denotes the plane at $z=1$ in the camera coordinate system. 
The red, green, and blue virtual sensors, corresponding to different cropped intrinsics, are placed on the planes at $z=f_1$ and $z=f_2$. The focal lengths $f_1$ and $f_2$, measured in meters, correspond to two effective focal lengths measured in pixel units. The projected areas of these three virtual sensors on the plane at $z=1$ are also denoted in red, green, and blue. 
$(u_i, v_i)$ denotes a pixel coordinate on the virtual sensor, and the corresponding 3D point projected onto the $z=1$ plane is $(x_i, y_i, 1)$.
For each virtual sensor, the corresponding projected area on the plane at $z=1$ geometrically encodes its unique view frustum. Consequently, we use the projected area as the encoding result of the cropped intrinsics.
}
\vspace{-0.2cm}
\label{fig:pe}
\end{figure}

\begin{figure}[t]
\centering
\includegraphics[width=1.0\linewidth]{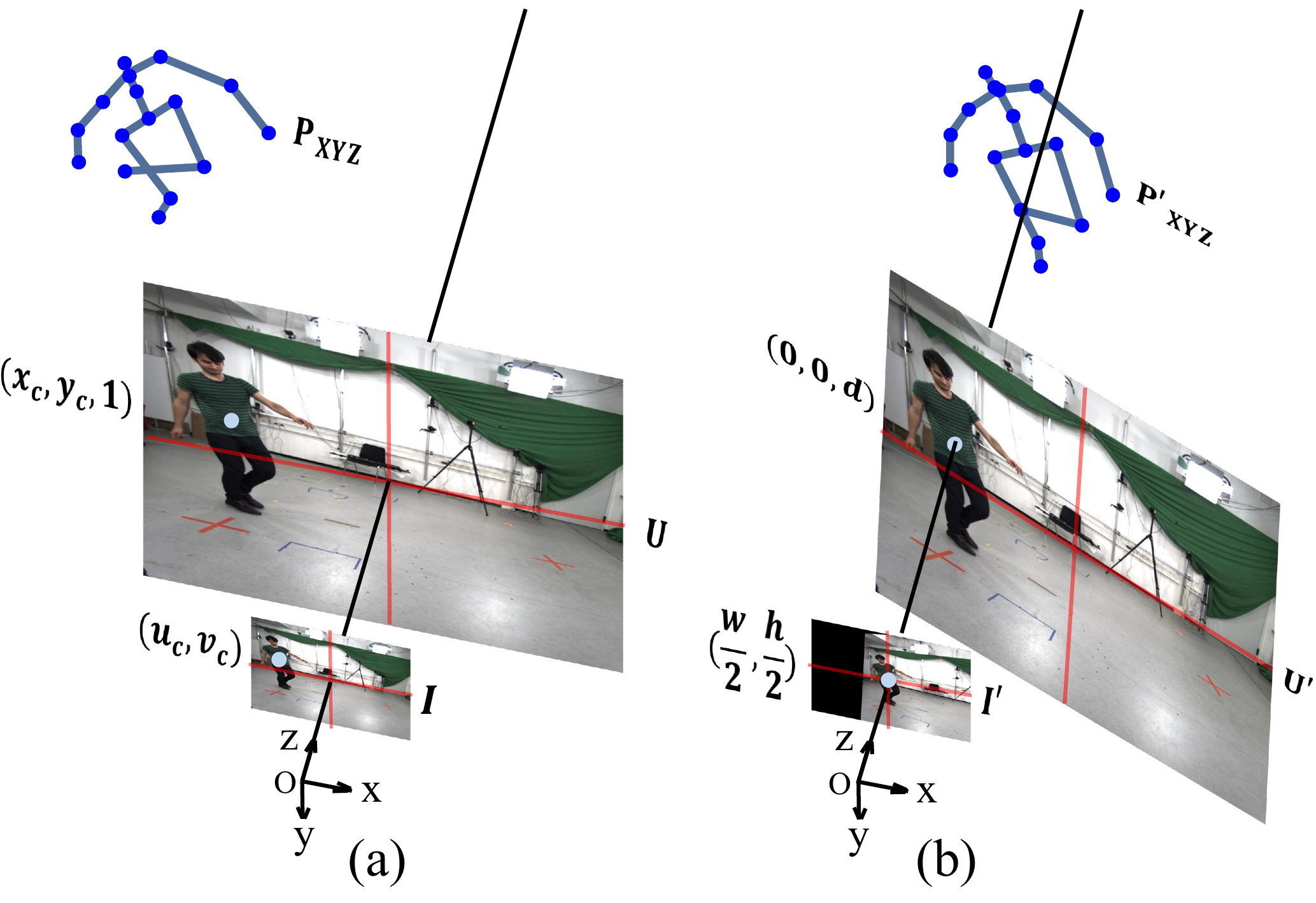}
\vspace{-0.6cm}
\caption{\textbf{Perspective Rotation}. In Figure (a), a virtual sensor captures an image $I$ that is marked with the human bounding box center $(u_c, v_c)$. An upscaled image $U$ is added to the scene, placed on the plane at $z = 1$, and the human bounding box center $(x_c, y_c, 1)$ on $U$ is also marked. Additionally, a 3D human skeleton $P_{\text{XYZ}}$ is added to the scene. 
The scene, including the 3D skeleton $P_{\text{XYZ}}$ and the upscaled image $U$, is rotated around the optical center $O$ from the setting in (a) to the setting in (b), so that the optical axis points to the human subject. After applying this rotation, the upscaled image $U^{\prime}$ in (b) is then reprojected onto the virtual sensor in (b), and the centered image $I^{\prime}$ in Figure \ref{fig:pipeline} is obtained.}
\vspace{-0.2cm}
\label{fig:pr} 
\end{figure}

In this section, we detail the proposed PersPose, illustrated in Figure \ref{fig:pipeline}. 
First, the original image $I$ undergoes the PR process, during which a rotation matrix $R$ is applied to center the human subject, yielding a centered image $I^{\prime}$ (see Section~\ref{sec:pr} for details). 
Then, the image $I^{\prime}$ is cropped from the center, resulting in the cropped image $I^{\text{crop}}$. 
Subsequently, the cropped intrinsics $K^{\text{crop}}$, calculated using Eq. \ref{eq:K_crop}, is encoded to form a 2D map $M^{xy}$ using PE, as detailed in Section~\ref{sec:encoding}. 
This PE map $M^{xy}$ and the cropped image $I^{\text{crop}}$ are separately processed through distinct convolutional layers before being element-wise added. The fused features are further passed through a backbone network, and then a decoder is used to predict 2D joint coordinates and relative depths of joints $P_{\text{UVD}}$, as well as a scale factor $\hat{s}=d^{\text{abs}}_0 / f^{\text{crop}}$. Here, $d^{\text{abs}}_0$ is the absolute depth of the pelvis joint in the camera coordinate system, and scale factor $\hat{s}$ is expressed in mm/pixel.
These outputs are then used to compute the rotated 3D pose $P^{\prime}_{\text{XYZ}}$. 
For a joint index by $i$, the calculation from $P_{\text{UVD},i}=[u_i,v_i,d_i]^\top$ to $P_{\text{XYZ},i}^{\prime}=[x_i^{\prime},y_i^{\prime},z_i^{\prime}]^\top$ is given by 
\begin{equation}
\begin{bmatrix}
x_i^{\prime} \\
y_i^{\prime} \\
z_i^{\prime}
\end{bmatrix}
= d_i^{\text{abs}}\,
(K^{\text{crop}})^{-1} 
\begin{bmatrix} u_i \\ v_i \\ 1 \end{bmatrix}, 
\end{equation}
where $d_i^{\text{abs}} = d_i + \hat{s}\,f^{\text{crop}}$.
Finally, the inverse rotation $R^\top$ is applied to $P^{\prime}_{\text{XYZ}}$ to derive $P_{\text{XYZ}}$, the 3D pose for the original image $I$.
\begin{equation}
P_{\text{XYZ}} = R^{\top}P^{\prime}_{\text{XYZ}}.
\end{equation}  

\subsection{Perspective Encoding}
\label{sec:encoding}

We propose PE to encode cropped intrinsics. As illustrated in Figure~\ref{fig:pe}, three virtual sensors have different cropped intrinsics $K^{\text{crop}}_{\text{red}}$, $K^{\text{crop}}_{\text{green}}$ and $K^{\text{crop}}_{\text{blue}}$. 
The red virtual sensor has a larger focal length compared to the green and blue virtual sensors.
Additionally, the red and green virtual sensors are aligned along the optical axis, while the blue virtual sensor is placed off-axis, which is common for cropped images. 
In this setup, the principal points of \(K^{\text{crop}}_{\text{red}}\) and \(K^{\text{crop}}_{\text{green}}\) are located at the image center, while the principal point of \(K^{\text{crop}}_{\text{blue}}\) is not. 
We project virtual sensors at different positions onto a fixed reference plane at $z=1$.
The red virtual sensor has a larger focal length than the other two virtual sensors, thus its corresponding projected region is smaller. Similarly, the principal point of the blue virtual sensor deviates from the optical axis, and its corresponding projected region also shifts away from the axis. In other words, for each sensor, the corresponding projected area geometrically represents the unique view frustum determined by its cropped intrinsics. Consequently, we employ this projected area as the encoded representation of the cropped intrinsics.

Specifically, we construct the PE map $M^{xy}$ by projecting each pixel coordinates $(u_i, v_i)$ of the cropped image onto the plane at \(z=1\), as shown in Figure~\ref{fig:pe}, and the projected coordinates are computed as:
\begin{equation}
\label{eq:xy1}
\begin{bmatrix}
x_i \\
y_i \\
1
\end{bmatrix} = 
\bigl(K^{\text{crop}}\bigr)^{-1}
\begin{bmatrix}
u_i \\
v_i \\
1
\end{bmatrix}.
\end{equation}
Consequently, the pixel coordinate map \(M^{uv}\), defined as
\begin{equation}
M^{uv} = 
\begin{bmatrix}
(1,1) & (1,2) & \cdots & (1,w) \\
(2,1) & (2,2) & \cdots & (2,w) \\
\vdots & \vdots & \ddots & \vdots \\
(h,1) & (h,2) & \cdots & (h,w) \\
\end{bmatrix},
\end{equation}
is transformed into the PE map \(M^{xy}\):
\begin{equation}
M^{xy} = 
\begin{bmatrix}
(x_1,y_1) & (x_1,y_2) & \cdots & (x_1,y_w) \\
(x_2,y_1) & (x_2,y_2) & \cdots & (x_2,y_w) \\
\vdots & \vdots & \ddots & \vdots \\
(x_h,y_1) & (x_h,y_2) & \cdots & (x_h,y_w) \\
\end{bmatrix}.
\end{equation}
The PE map $M^{xy}$ encodes the view frustum and is subsequently fed into the 3D HPE model, enabling the model to capture the perspective variations induced by different cropped intrinsics.

\subsection{Perspective Rotation}
\label{sec:pr}
In Section \ref{sec:importance}, we explain the importance of camera intrinsics in 3D HPE. Then, we introduce PE to encode cropped intrinsics $K^{\text{crop}}$ as part of model input in Section \ref{sec:encoding}. 
Given a cropped image $I^{\text{crop}}$ and the corresponding cropped intrinsics $K^{\text{crop}}$ (composed of $f^{\text{crop}}$, $c_x^{\text{crop}}$, and $c_y^{\text{crop}}$), a 3D HPE model, parameterized by $\theta$, aims to learn the following mapping function:
\begin{equation}
\label{eq:mapping_before}
f_{\theta}: (I^{\text{crop}}, f^{\text{crop}}, c_x^{\text{crop}}, c_y^{\text{crop}}) \rightarrow P_{\text{XYZ}}. 
\end{equation}
Notably, the principal point $(c_x^{\text{crop}}, c_y^{\text{crop}})$ may exhibit considerable variance across different images, as the human subject can appear at any region within the original image. This substantially increases the difficulty of model fitting. 
To mitigate this challenge, we introduce PR to center the human subject, as shown in Figure \ref{fig:pipeline}. 
This ensures that the cropped principal point $(c_x^{\text{crop}}, c_y^{\text{crop}})$ is fixed to the center of the cropped image across different images.
Therefore, the mapping function changes from Eq. \ref{eq:mapping_before} to
\begin{equation}
\label{eq:mapping_after}
\tilde{f}_{\theta}: (I^{\text{crop}}, f^{\text{crop}}) \rightarrow P_{\text{XYZ}},
\end{equation}
thus reducing the model's fitting difficulty.

In the following, we describe the process of PR. As illustrated in Figure \ref{fig:pipeline}, through PR, the human subject in the original image $I$ is reprojected to the center in the image $I^{\prime}$. The reprojection process of this example is shown in Figure \ref{fig:pr}.

In Figure \ref{fig:pr}(a), the virtual sensor, the upscaled image $U$, and the 3D skeleton $P_{\text{XYZ}}$ form a perspective relationship: if a ray is projected from a particular joint of the $P_{\text{XYZ}}$ towards the optical center $O$, this ray passes through the corresponding joint in the original image $I$ on the sensor, and also through the corresponding joint in the upscaled image $U$.

The coordinates $(u_c, v_c)$ and $(x_c, y_c, 1)$ in Figure \ref{fig:pr}(a) denote the center of the human bounding box on the sensor and on the upscaled image $U$, respectively. 
Here, $(x_c, y_c, 1)$ is calculated as 
\begin{equation}
\begin{bmatrix}
x_c \\
y_c \\
1
\end{bmatrix} =
K^{-1}
\begin{bmatrix}
u_c \\
v_c \\
1
\end{bmatrix},
\end{equation}
where $K$ is the camera intrinsics of the original image $I$.

The goal of PR is to center the bounding box in the centered image $I^{\prime}$. 
To achieve this, we rotate the $U$ around the optical center such that the bounding box center $(x_c, y_c, 1)$ is brought onto the $z$-axis at $(0, 0, d)$, where $d = \|(x_c, y_c, 1)\|_2$, as shown in Figure \ref{fig:pr}(b). 
Therefore, the rotation axis $\mathbf{n}$ and rotation angle $\phi$ can be computed as follows: 
\begin{align}
\mathbf{n} &= \frac{(x_c, y_c, 1) \times (0,0,d)^\top}{\|(x_c, y_c, 1) \times (0,0,d)^\top\|}, \\
\phi &= \arccos \left(\frac{\left(x_{c}, y_{c}, 1\right) \cdot(0,0, d)}{\left\|\left(x_{c}, y_{c}, 1\right)\right\| \|(0,0, d)\|}\right),
\end{align}
where $\times$ denotes the cross product, and  $\cdot$ represents the dot product.
Applying Rodrigues' rotation formula, the rotation matrix is:
\begin{equation}
R = \text{Rodrigues}(\mathbf{n}, \phi)
\end{equation}

This rotation $R$ is first applied to the upscaled image $U$ in Figure \ref{fig:pr}(a) and obtain $U^{\prime}$ in Figure \ref{fig:pr}(b). $U^{\prime}$ is then reprojected onto the virtual sensor in (b) and the centered image $I^{\prime}$ is obtained, with the bounding box center located at image center $\left(\frac{w}{2}, \frac{h}{2}\right)$.

The 3D pose corresponding to the centered image $I^{\prime}$ can be derived by applying the same rotation $R$ to $P_{\text{XYZ}}$:
\begin{equation}
\label{eq:p'}
P'_{\text{XYZ}} = RP_{\text{XYZ}}.
\end{equation}  
Since both $U$ and $P_{\text{XYZ}}$ undergo the same 3D rotation about the optical center, the perspective relationship of the centered image $I^{\prime}$, the rotated upscaled image $U^{\prime}$, and the rotated 3D skeleton $P'_{\text{XYZ}}$ is maintained, as illustrated in Figure \ref{fig:pr}(b). Consequently, 
the rotated 3D skeleton $P'_{\text{XYZ}}$ maintains geometric consistency with the new image $I^{\prime}$.

Finally, the perspective transformation matrix $M$ used to warp the original image $I$ into the centered image $I^{\prime}$ is computed as follows:
\begin{equation}
M = K R K^{-1},
\end{equation}
where $K$ is the camera intrinsics of the original image $I$.

\section{Experiments}

\setlength{\tabcolsep}{2.5pt} 
\begin{table*} [t]
\centering
\begin{tabular}{cc|ccc|ccc|ccc}
\toprule
 & & \multicolumn{3}{c|}{3DPW} & \multicolumn{3}{c|}{Human3.6M}  & \multicolumn{3}{c}{MPI-INF-3DHP} \\
\cline{3-11}
 & & PA-MPJPE${\downarrow}$ & MPJPE${\downarrow}$ & PVE${\downarrow}$ & PA-MPJPE${\downarrow}$ & MPJPE${\downarrow}$ & PVE${\downarrow}$ & PCK${\uparrow}$ & AUC${\uparrow}$ & MPJPE${\downarrow}$ \\
 \cline{1-11}
HMR   \cite{kanazawa2018hmr} $^\dagger$ & CVPR'18  & 81.3          & 130.0         & 152.7         & 56.8          & 88.0            & 96.1        & 72.9          & 36.5          & 124.2         \\
SPIN \cite{SPIN2019iccv} $^\dagger$      & ICCV'19  & 59.2          & 96.9        & 116.4         & 41.1          & -             & -           & 76.4          & 37.1          & 105.2         \\
I2L-MeshNet \cite{i2l-meshnet} $^\dagger$     & ECCV'20  & 57.7          & 93.2        & 110.1         & 41.1          & 55.7          & 65.1        & -             & -             & -             \\
Pose2Mesh \cite{choi2020pose2mesh}   $^\dagger$  & ECCV’20  & 58.3          & 88.9        & 106.3         & 46.3          & 64.9          & 85.3        & -             & -             & -             \\
Mesh Graphormer \cite{lin2021mesh}   & ICCV'21  & 45.6          & 74.7        & 87.7          & 34.5          & 51.2          & -           & -             & -             & -             \\
HybrIK \cite{HybrIK2021cvpr} $^\ddagger$    & CVPR'21  & 45.0            & 74.1        & 86.5          & 34.5          & 54.4          & 65.7        & 87.5          & 46.9          & 93.9          \\
CLIFF \cite{li2022cliff}    & ECCV'22  & 43.0            & 69.0          & 81.2          & 32.7          & 47.1          & -           & -             & -             & -             \\
FastMETRO \cite{cho2022cross}    & ECCV'22  & 44.6          & 73.5        & 84.1          & 33.7          & 52.2          & -           & -             & -             & -             \\
IKOL \cite{zhang2023ikol} $^\ddagger$        & AAAI'23  & 45.5          & 73.3        & 86.4          & -             & -             & -           & 87.9          & 48.1          & 88.8          \\
VirtualMarker \cite{ma20233d} $^\ddagger$    & CVPR'23  & 41.3          & 67.5        & 77.9          & 32.0            & 47.3          & \underline{58.0} & -    & -    & -    \\
ProPose   \cite{fang2023learning} $^\ddagger$ & CVPR'23  & 40.6          & 68.3        & 79.4          & \underline{29.1} & \underline{45.7} & -           & -             & -             & -             \\
PLIKS \cite{shetty2023pliks} $^\ddagger$      & CVPR'23  & 42.8          & 66.9        & 82.6          & 34.7          & 49.3          & -           & \underline{91.8} & \underline{52.3} & \underline{72.3} \\
Zolly \cite{wang2023zolly}                      & ICCV'23  & 39.8          & \underline{65.0} & \underline{76.3} & 32.3          & 49.4          & -           & -             & -             & -        \\
Gwon et al. \cite{gwon2024instance}                & CVPR'24  & 44.3          & 73.2        & 80.3          & -             & -             & -           & -             & -             & -             \\
GLNet-W48 \cite{xiao2024global}                    & ECCV'24  & \underline{39.5} & 66.9        & 77.9          & 29.4          & 48.8          & -           & -             & -             & -             \\
PostoMETRO \cite{yang2024postometro}               & WACV'25  & 39.8          & 67.7        & 76.8          & -             & -             & -           & -             & -             & -             \\
\hline
\textbf{PersPose $^\ddagger$} &  & \textbf{39.1} & \textbf{60.1} &  \textbf{72.4} &  \textbf{28.3} & \textbf{43.0} & \textbf{52.7} & \textbf{94.0} & \textbf{55.2} &  \textbf{72.1} \\
\bottomrule
\end{tabular} 
\vspace{-0.1cm}
\caption{\textbf{Comparison between our method and SOTA HPS methods}. $\downarrow$ indicate that lower values are better, while $\uparrow$ indicate that higher values are better. $\dagger$ denotes that 3DPW training split is not used. $\ddagger$ denotes methods that estimate UVD coordinates as an intermediate representation prior to obtaining 3D XYZ positions. }
\vspace{-0.3cm} 
\label{tab:mesh}
\end{table*}
\setlength{\tabcolsep}{6pt} 

\subsection{Experimental setup}
We set the resolution of the cropped images to $256 \times 256$. The HRNet-W48~\cite{sun2019deep} is used as the backbone, which outputs a feature map of dimensions $64 \times 64 \times 48$ and a global feature vector of length $2048$. 
The Perspose decoder comprises one convolutional layer and two linear layers. The convolutional layer processes the feature map to produce a 2D heatmap, from which the 2D pose is obtained using the soft-argmax operation. Simultaneously, the two linear layers take the global feature vector as input to predict the relative depths of joints and the scale factor $\hat{s}$, respectively. 
Additionally, the element-wise addition in Figure~\ref{fig:pipeline} is performed before the second stage of HRNet. 

To compare with HPS methods, 
we additionally add two linear layers into the Perspose decoder to estimate human shape and twist, and the analytical IK algorithm in HybrIK~\cite{HybrIK2021cvpr} is used to derive SMPL parameters prediction. 

The loss function is defined as 
\begin{equation}
    \mathcal{L} = \lambda_1 \mathcal{L}_{\text{uvd}} + \lambda_2 \mathcal{L}_{\hat{s}} + \lambda_3 \mathcal{L}_{\text{beta}} + \lambda_4 \mathcal{L}_{\text{tw}},
\end{equation}
where $\mathcal{L}_{\text{uvd}}$ is the L1 loss for the $P_{\text{UVD}}$ predictions, $\mathcal{L}_{\hat{s}}$ is the L1 loss for the estimated scale factor $\hat{s}$, 
$\mathcal{L}_{\text{beta}}$ is the L1 loss for the estimated human shape parameters,
$\mathcal{L}_{\text{tw}}$ is the L2 loss for the twist predictions, 
and $\lambda_1$, $\lambda_2$, $\lambda_3$ and $\lambda_4$ are weight hyperparameters, set to 1.0, 0.5, 0.01 and 0.1 respectively.

The batch size is set to 96. The AdamW~\cite{loshchilov2017decoupled} optimizer is used with an initial learning rate of 3e-4. The training lasts for 70 epochs, and the learning rate is reduced by a factor of 0.75 every 6 epochs. This process utilizes one NVIDIA 3090 GPU and is completed in around 60 hours.

To evaluate the estimated 3D pose, we report Mean Per Joint Position Error (MPJPE), Procrustes Aligned MPJPE (PA-MPJPE), Percentage of Correct Keypoints (PCK), and Area Under Curve (AUC). 
Additionally, Per Vertex Error (PVE) is computed to evaluate all vertices on the human mesh. Furthermore, we report on the depth error in our ablation experiments to evaluate the estimated relative depths. The depth error is defined as the mean absolute difference between the estimated and ground-truth relative depths of joints.  Following~\cite{HybrIK2021cvpr,ma20233d}, We evaluate 14 joints on 3DPW and Human3.6M datasets and 17 joints on MPI-INF-3DHP dataset. 
For MPJPE, PA-MPJPE, PCK, and AUC, the joints regressed from the human mesh are evaluated, while the relative depths predicted by the PersPose decoder is used to compute depth error. 

Our experiments employ the following 3D datasets: 
1) 3DPW~\cite{3DPW2018}, an in-the-wild dataset annotated with SMPL parameters, which can be used to derive 3D pose labels. 
2) Human3.6M~\cite{human36M2013}, which was captured in a controlled environment. 
We use the SMPL parameters derived from Mosh~\cite{loper2014mosh}. 
Following~\cite{kanazawa2019hmmr,VIBE2020cvpr}, we use 5 subjects (S1, S5, S6, S7, S8) for training, and 2 subjects (S9, S11) for evaluation. 
3) BEDLAM~\cite{black2023bedlam}, a large and realistic synthetic dataset annoated with accurate SMPL parameters. 
4) MPI-INF-3DHP~\cite{mehta2017monocular}, a 3D human pose dataset that includes both controlled indoor settings and complex outdoor scenes.

\subsection{Comparison with the State-of-the-art}
Table~\ref{tab:mesh} presents a comparison between our method and SOTA HPS methods on 3DPW, Human3.6M, and MPI-INF-3DHP datasets. 
In this study, we employ HRNet-W48 \cite{sun2019deep} as the backbone, 
and results utilizing the same backbone from other works will be preferred when available.
The results for the proposed PersPose in Table~\ref{tab:mesh} were obtained by training on a combination of datasets, including 3DPW~\cite{3DPW2018}, Human3.6M~\cite{human36M2013}, MPI-INF-3DHP~\cite{mehta2017monocular}, and COCO~\cite{lin2014microsoft}. 
Note that while some methods may have incorporated additional 2D keypoint datasets during training, none of the results in Table~\ref{tab:mesh} were derived using synthetic 3D datasets.

PersPose outperforms other methods across all three datasets. On MPI-INF-3DHP benchmark, PersPose surpasses the second-best approach by 2.2 on PCK and 2.9 on AUC. 
Furthermore, on 3DPW dataset, PersPose outperforms the second-best approach by 4.9 mm (7.54\%) in MPJPE. 
The test splits of MPI-INF-3DHP and 3DPW datasets both contain complex outdoor scenes, demonstrating the robustness of the proposed PersPose under challenging real-world conditions.

\subsection{Ablation Study}

\begin{table*}[ht]
\centering
\begin{tabularx}{\textwidth}{C{0.05\textwidth}C{0.05\textwidth}|A|C{0.11\textwidth}C{0.11\textwidth}A|C{0.11\textwidth}C{0.11\textwidth}A}
\toprule
\multirow{2}{*}{PR} & \multirow{2}{*}{PE} & \multirow{2}{*}{Dataset} & \multicolumn{3}{c|}{3DPW}       & \multicolumn{3}{c}{MPI-INF-3DHP} \\
\cline{4-9}
                    &                     &                          & Depth error${\downarrow}$ & PA-MPJPE${\downarrow}$ & MPJPE${\downarrow}$ & Depth error${\downarrow}$  & PA-MPJPE${\downarrow}$  & MPJPE${\downarrow}$ \\
\hline
-                   & -                   & \multirow{3}{*}{R}                        & 45.1        & 39.8     & 62.4  & 57.3         & 57.3      & 80.1  \\
-                   & \checkmark                   &                          & 44.5        & 39.7     & 62.2  & 53.7         & 56.3      & 76.6  \\
\checkmark                   & \checkmark                   &                          & \textbf{43.8}        & \textbf{39.1}     & \textbf{60.1}  & \textbf{51.0}           & \textbf{54.4}      & \textbf{71.9}  \\
\hline
-                   & -                   & \multirow{3}{*}{R+B}                       & 41.5        & 37.8     & 58.4  & 54.2         & 55.5      & 76.8  \\
-                   & \checkmark                   &                          & 41.2        & 37.8     & 58.1  & 51.0           & 55.1      & 73.4    \\
\checkmark                   & \checkmark                   &                          & \textbf{40.0}          & \textbf{37.3}     & \textbf{57.2}  & \textbf{48.6}    & \textbf{53.7}      & \textbf{70.2} \\
\bottomrule
\end{tabularx}
\vspace{-0.1cm}
\caption{\textbf{Ablation experiments on 3DPW~\cite{3DPW2018} and  MPI-INF-3DHP~\cite{mehta2017monocular}}. R denotes the use of real datasets, and B denotes BEDLAM.}
\vspace{-0.3cm}
\label{tab:abl}
\end{table*}

\paragraph{Experimental Setup.}
To evaluate the effectiveness of our PE and PR modules, we conduct ablation studies comprising the following configurations: 1) our framework without PE and PR, 2) our framework without PR, and 3) our framework with both PE and PR. 

For evaluation metrics, we employ PA-MPJPE and MPJPE to assess the accuracy of 3D pose estimation. 
In addition, Section \ref{sec:importance} highlights the importance of camera intrinsics for relative depth estimation with two examples (Figure \ref{fig:crop1} and Figure \ref{fig:crop2}). Section \ref{sec:encoding} introduces PE to transfer camera intrinsics to the neural model. Section \ref{sec:pr} proposes PR to reduce the difficulty of model fitting. 
To evaluate the effectiveness of our proposed PE and PR in estimating relative depths, we additionally report on the depth error. 

Prior work has demonstrated that training on synthetic data improves accuracy \cite{black2023bedlam,shin2023wham,shetty2023pliks}. In this study, we further investigate the impact of incorporating synthetic data into the training phase. Specifically, in Table~\ref{tab:abl}, ``R'' represents training on real-image datasets (3DPW, Human3.6M, MPI-INF-3DHP, and COCO), while ``B'' denotes training on BEDLAM synthetic dataset.

\paragraph{Ablation Study Results.}
Table~\ref{tab:abl} displays the results of our experiments. The addition of PE or PR consistently leads to performance improvements across both datasets, regardless of whether the additional synthetic dataset is used for training, indicating the effectiveness of the proposed PE and PR in relative depth estimation and 3D pose estimation. 

On MPI-INF-3DHP dataset, the addition of PE or PR results in a considerable reduction in depth error. Specifically, PE reduces depth error by 3.6mm and 3.2mm for training settings ``R'' and ``R+B'', respectively. Similarly, PR achieves reductions in depth error of 2.7mm and 2.4mm for ``R'' and ``R+B'', respectively.

On 3DPW dataset, incorporating PE or PR leads to a relatively modest performance improvement compared to MPI-INF-3DHP dataset. To investigate this effect, we analyzed the distribution of the rotation angle $\phi$ in PR across different datasets. 

\begin{figure}[t]
  \centering
   \includegraphics[width=0.65\linewidth]{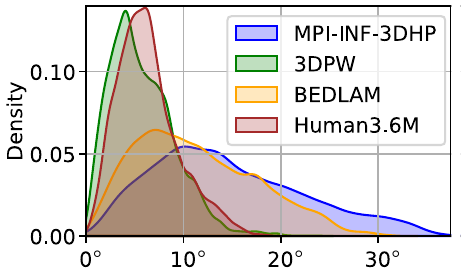}
  \vspace{-0.2cm}
  \caption{\textbf{Distribution of PR rotation angle $\phi$}. We analyze the distribution of $\phi$ (measured in degrees) across various datasets. A broader spread in $\phi$ reflects greater variability in the centre of the bounding box's coordinates $(u_c, v_c)$. The values of $\phi$ for 3DPW and Human3.6M concentrate around $5^{\circ}$. In contrast, BEDLAM and MPI-INF-3DHP exhibit a wider spread, suggesting these datasets feature more diversity and complexity. 
  } 
   \label{fig:dist}
   \vspace{-0.4cm}
\end{figure}

\paragraph{Distribution of the rotation angle $\phi$ across datasets.}
Figure~\ref{fig:dist} displays the distribution of $\phi$, using kernel density estimation to estimate the density curve. The rotation angle $\phi$ for MPI-INF-3DHP spans a broader range than 3DPW. 
This distribution quantifies the variability in the person's position within the dataset's images: a wider spread in $\phi$ indicates greater variation in the center of the human bounding box $(u_c, v_c)$, 
reflecting significant changes in the perspective relationship between the 3D scene and the cropped image; in such cases, the benefits of PE and PR become even more pronounced.

\paragraph{Effect of Synthetic Training Data.}
Comparing the training settings ``R+B'' and ``R'' across various model configurations, we observe that the model trained with ``R+B'' consistently outperforms the one trained with ``R'' under identical model configurations. 
For instance, by incorporating additional synthetic training data, the MPJPE of our framework with PE and PR decreases from 60.1 mm to 57.2 mm on 3DPW dataset and from 71.9 mm to 70.2 mm on MPI-INF-3DHP dataset. 
This underscores the beneficial impact of incorporating additional synthetic datasets during training.

\section{Conclusion}
In this paper, we propose a novel 3D HPE framework, PersPose, integrating PE and PR. 
We first underscore the necessity of camera intrinsics for accurately inferring the relative depths of joints in 3D HPE and then introduce the PE module to encode camera intrinsics as a 2D map, which is jointly fed into the CNN alongside the cropped image. 
Considering that the human subject may appear anywhere in an image, we propose the PR module to center the human subject, thereby reducing perspective distortion and the difficulty of model fitting. 
Experimental results on the multiple datasets demonstrate that PersPose achieves considerable improvement over existing SOTA methods. 
Ablation studies confirm the effectiveness of PE and PR and demonstrate that the performance enhancements provided by these modules become more pronounced when perspective relationships are more variable.

{
    \small
    \bibliographystyle{ieeenat_fullname}
    \bibliography{main}
}

\end{document}